\pdfoutput=1
\documentclass[10pt,twocolumn,letterpaper]{article}

\usepackage[pagenumbers]{cvpr} 

\usepackage{graphicx}
\usepackage{amsmath}
\usepackage{amssymb}
\usepackage{booktabs}

\usepackage{xcolor}
\usepackage{multirow}
\usepackage{stfloats}
\usepackage{adjustbox}

\usepackage[pagebackref,breaklinks,colorlinks]{hyperref}

\usepackage[capitalize]{cleveref}
\crefname{section}{Sec.}{Secs.}
\Crefname{section}{Section}{Sections}
\Crefname{table}{Table}{Tables}
\crefname{table}{Tab.}{Tabs.}

\def\cvprPaperID{1722} 
\def\confName{CVPR}
\def\confYear{2023}

\definecolor{myGreen}{HTML}{559D64}
\definecolor{myBlue}{HTML}{3D69B9}

\newcommand*\samethanks[1][\value{footnote}]{\footnotemark[#1]}
\newcommand{\ins}[1]{$^{#1}$}

\begin{document}

\title{Propagate And Calibrate: Real-time Passive Non-line-of-sight Tracking}

\author{Yihao Wang\ins{1}\thanks{Equal contribution.} \quad Zhigang Wang\ins{1}\samethanks[1] \quad Bin Zhao\ins{1,2}\thanks{Corresponding author.} \quad Dong Wang\ins{1} \quad Mulin Chen\ins{1,2} \quad Xuelong Li\ins{1,2}\samethanks[2]\\
\ins{1}Shanghai AI Laboratory \qquad \ins{2}Northwestern Polytechnical University\\
{\tt\small \{wangyihao,wangzhigang,zhaobin,wangdong,chenmulin\}@pjlab.org.cn \qquad li@nwpu.edu.cn}
}
\maketitle

\begin{abstract}
    \vspace{-3mm}
    Non-line-of-sight (NLOS) tracking has drawn increasing attention in recent years, due to its ability to detect object motion out of sight. Most previous works on NLOS tracking rely on active illumination, \eg, laser, and suffer from high cost and elaborate experimental conditions. Besides, these techniques are still far from practical application due to oversimplified settings. In contrast, we propose a purely passive method to track a person walking in an invisible room by only observing a relay wall, which is more in line with real application scenarios, \eg, security. To excavate imperceptible changes in videos of the relay wall, we introduce difference frames as an essential carrier of temporal-local motion messages. In addition, we propose PAC-Net, which consists of alternating propagation and calibration, making it capable of leveraging both dynamic and static messages on a frame-level granularity. To evaluate the proposed method, we build and publish the first dynamic passive NLOS tracking dataset, NLOS-Track, which fills the vacuum of realistic NLOS datasets. NLOS-Track contains thousands of NLOS video clips and corresponding trajectories. Both real-shot and synthetic data are included. Our codes and dataset are available at \href{https://againstentropy.github.io/NLOS-Track/}{https://againstentropy.github.io/NLOS-Track/}.
    \looseness=-1
\end{abstract}

\vspace{-5mm}
\section{Introduction}
\label{sec:intro}

\begin{figure}[t]
    \centering
    \includegraphics[width=\linewidth]{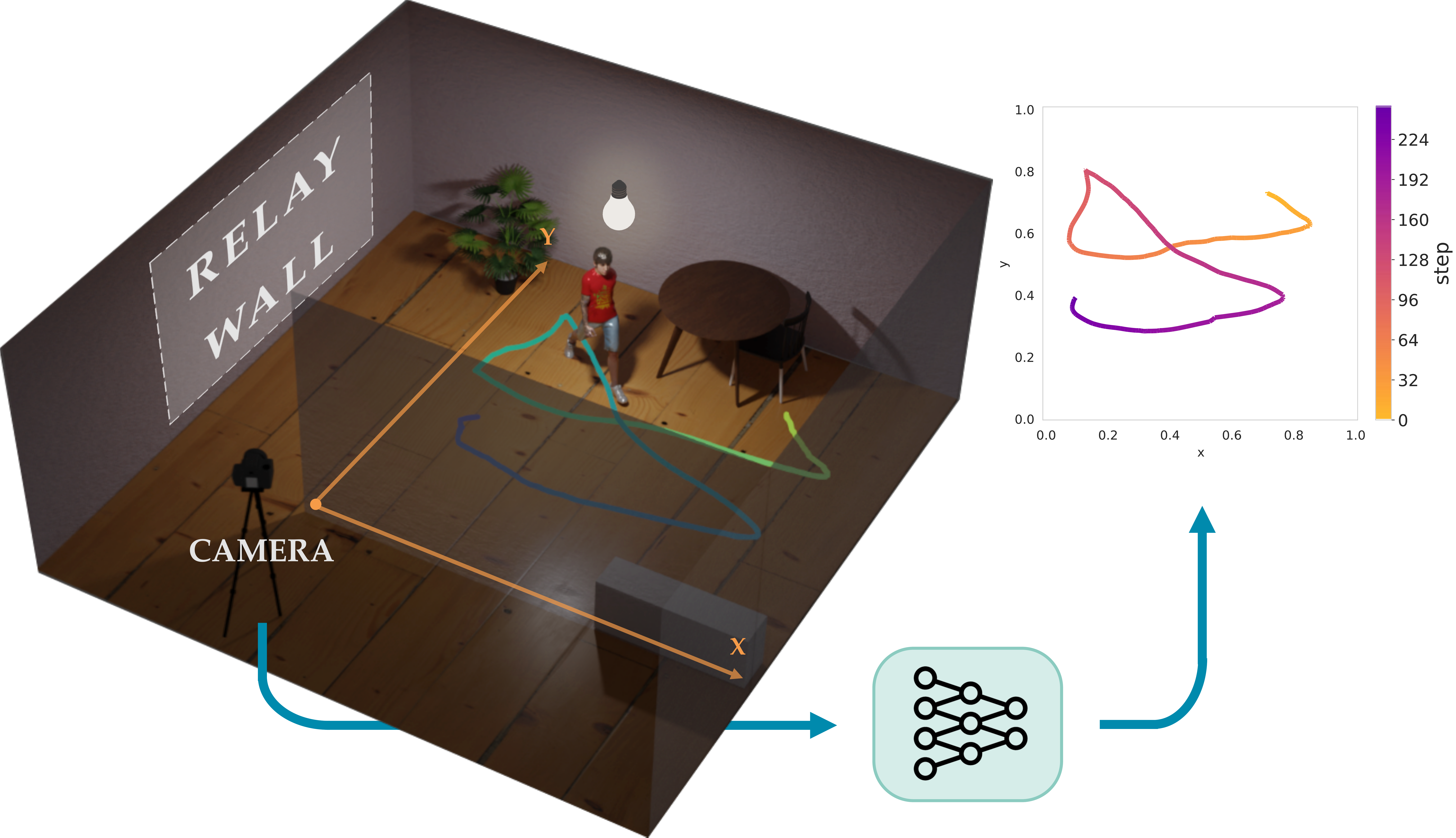}
    \caption{\textbf{A schematic of the passive NLOS tracking.} The character is walking in the hidden scene and we can perform real-time tracking by observing and analyzing the relay wall from outside the room with a RGB camera, without any additional equipment.
    \looseness=-1}
    \label{fig:teaser}
    \vspace{-10pt}
\end{figure}

In contrast to conventional imaging within the direct line-of-sight (LOS), non-line-of-sight (NLOS) imaging aims to tackle an inverse problem, \ie, using indirect signal (\eg, reflection from a visible relay wall) to recover information of invisible areas. To specify, NLOS tracking manages to reconstruct a continuous trajectory in real time when an object or a person is moving in an invisible region, which is sketched in \cref{fig:teaser}.
The ability to track moving objects outside the LOS would enable promising applications, such as autonomous driving, robotic vision, security, medical imaging, post-disaster searching, and rescue operations, \etc \cite{Borges2012Pedestrian, maeda2019recent, faccio2020non, geng2021recent}, thus receiving increasing attention in recent years.

Existing NLOS tracking techniques mostly rely on active illumination from the detection side \cite{gariepy2016detection, klein2016tracking, klein2016transient, chan2017fast, Chan:17, smith2018tracking, tancik2018flash, brooks2019single, metzler2020keyhole, cao2022computational}. 
Although introducing denser and finer information, active illumination typically requires expensive equipment (\eg, ultra-fast pulsed laser) and elaborate experimental conditions \cite{saunders2019computational}. 
These defects cause a gap between active techniques and practical applications. Besides, the oversimplified setting in previous works even expands the gap. Unlike active methods, passive NLOS techniques \cite{baradad2018inferring, saunders2019computational, yedidia2019using, wang2021accurate, geng2022passive, seidel2019corner, Seidel2021TwoDim, bouman2017turning, PrafullSharma2021WhatYC, he2022non, cao2022computational} only depend on the feeble diffuse reflection of the hidden region, getting rid of requirements of expensive equipment. So this paper focuses on the low-cost passive NLOS tracking task in realistic scenarios. 
\looseness=-1

We find that most existing NLOS tracking works merely locate the object in each frame independently \cite{klein2016transient, chan2017fast, Chan:17, tancik2018flash, brooks2019single, metzler2020keyhole, cao2022computational, he2022non}, without considering the position relationship between adjoining moments. This practice directly causes jitters of trajectory, thus resulting in inaccurate tracking (see \cref{sec:c-net} for more details). In this paper, we consider the significance of making use of motion information and taking advantage of motion continuity prior, which helps achieve more coherent and accurate tracking results.

Furthermore, passive NLOS techniques face the dilemma that the signal-to-noise ratio (SNR) is extremely low \cite{Chan:17, caramazza_2018}. To address this problem, some previous works conduct background estimation with the video's temporal mean and apply background subtraction to every frame \cite{bouman2017turning, PrafullSharma2021WhatYC, he2022non}. In this way, the difference between frames could be amplified, thus increasing the SNR.
However, temporal-mean subtraction inevitably mixes up information from early period. Consequently, it reintroduces extra noise into originally low-SNR signals, which is still a hazard to excavating faint differences between frames.

To address the aforementioned problems, we first introduce \textit{difference frame} to describe motion information.
Compared to background estimation and subtraction, a difference frame can be readily obtained by subtracting the previous frame from the current frame.
In this way, a difference frame can represent the immediate motion information, and will not introduce noise from other periods. Our experiments show that difference frames do convey essential dynamic messages (see \cref{sec:p-net} for more details).
Additionally, we propose a novel network named PAC-Net (\textbf{P}ropagation \textbf{A}nd \textbf{C}alibration \textbf{Net}work), which integrates motion continuity prior into the algorithm. Consisting of two dual modules, Propagation-Cell and Calibration-Cell, PAC-Net maintains a good continuity of trajectory via propagating with difference frames and then alternately calibrating with raw frames. 
Our experimental results demonstrate that PAC-Net can achieve centimeter-level precision when tracking a walking person in real time.

We also build NLOS-Track, the first public-accessible video dataset for passive NLOS tracking. It contains realistic scenes to support the proposed task and method, and we expect NLOS-Track to facilitate more NLOS works. In contrast to oversimplified settings in existing NLOS tracking works, NLOS-Track dataset manages to simulate realistic scenarios with humans walking in unknown scenes. The dataset consists of 500 real-shot videos and more than 1,000 synthetic videos, each recording the relay wall when a character walks along the randomly generated trajectory. Paired trajectory ground truth of each video clip is also provided. 

Our contributions are mainly in three folds:
\begin{itemize}
    \item We propose and formulate the purely passive NLOS tracking task, which avoids the use of expensive equipment. Development on this task will allow promising and valuable applications in many fields, such as robotic vision, medical imaging, \etc.
    \item We propose a passive NLOS tracking network, PAC-Net, which is capable of utilizing both dynamic and static messages on a frame level. As for dynamic messages, we specially introduce difference frames as clear carriers of motion information, which gets rid of introducing extra noise from other periods.
    \item We establish the first passive NLOS trajectory tracking dataset, NLOS-Track, which contains thousands of video clips with a variety of scene settings.
\end{itemize}

\section{Related Work} \label{sec:related}

\noindent \textbf{Passive NLOS.}~
Previous passive NLOS imaging techniques mainly focus on reconstructing static information of the invisible scene. Some works leverage pinholes or pinspecks as ``accidental cameras" \cite{cohen1982anti, torralba2012accidental} while others rely on occluders (\eg, blocking objects or corners). These works make use of shadows and penumbrae cast on the visible wall or floor to extract useful information about the hidden scene \cite{baradad2018inferring, saunders2019computational, yedidia2019using, wang2021accurate, geng2022passive, seidel2019corner, Seidel2021TwoDim}. 

As for the dynamic NLOS scenario, it was first shown by Bouman \etal \cite{bouman2017turning} that obstructions with edges can be exploited as ``corner cameras". They reveal the number and trajectories of people moving in an occluded scene with recovered 1-D spatio-temporal videos.
Sharma \etal \cite{PrafullSharma2021WhatYC} presented a deep learning method that reveals the number or activity of people in an unknown room by observing a blank relay wall.
Wang \etal \cite{Wang2022event} first proposed a novel method for NLOS moving target reconstruction, which uses an event camera to extract rich dynamic information of the speckle movement. 

Compared to existing methods, our technique doesn't introduce any additional structures or special devices. With only a visible blank wall and a conventional RGB camera, we can extract both motion and static information on the frame level and perform tracking in real time.

\noindent \textbf{Active NLOS localization and tracking.}~ 
Some previous works accomplished NLOS tracking directly through locating the hidden object or person frame-by-frame~\cite{klein2016transient, chan2017fast, Chan:17, tancik2018flash, brooks2019single, metzler2020keyhole, cao2022computational}, whereas other methods consider object motion to assist tracking~\cite{gariepy2016detection, klein2016tracking, smith2018tracking}. All methods mentioned above rely on active illumination and most of them rely on time-resolved detection techniques. Methods employing lasers typically take advantage of the high flatness of optical experimental platforms. In contrast, our method removes the need for any additional illumination, equipment, and special detector. 
\looseness=-1

\noindent \textbf{NLOS datasets.}~ 
Large-scale, labeled and readily accessible datasets are vital to technique development. However, only few NLOS works provide datasets, and most of them focus on active NLOS imaging.
Jarabo \etal \cite{jarabo2014framework} proposed an effective framework for rendering in transient state, which has been exploited by several following works for data generation \cite{klein2018quantitative, liu2019phasor, zhu2022fast}.
Klein \etal \cite{klein2018quantitative} released a synthetic data foundation with a few scenes and the first reconstruction benchmark platform for a variety of NLOS imaging tasks, along with task-specific quality metrics. To further expand the data scale and facilitate data-driven methods, a new benchmark dataset for time-resolved NLOS imaging, Z-NLOS, is proposed by Galindo \etal \cite{galindo2019dataset}.

As for passive datasets, Chen \etal have presented the first large-scale static passive NLOS dataset\cite{geng2022passive}. Wang \etal \cite{Wang2022event} created the first event-based NLOS imaging dataset, which explores a novel modal in dynamic NLOS imaging. 
In addition to the fact that only a proportion of datasets are available, the lack of realistic dynamic passive NLOS datasets also remains an obstacle to exploring passive NLOS methods. To address this issue, we propose a new dataset, NLOS-Track, which contains both synthetic data and real-shot data.

\section{Problem Formulation and Signal Extraction}

\subsection{NLOS Tracking Problem}

Passive NLOS imaging aims to excavate information about a hidden scene through the diffuse reflection of ambient light. This task can be accomplished by observing and analyzing a relay wall. Every slight change in the hidden room could induce an imperceptible variation of the reflection, thus influencing the wall image.
The complicated optical system can be formulated as an imaging function $\mathcal{F}$:
\begin{equation} \label{eq:img_map}
\vspace{-1pt}
    I = \mathcal{F}(\Vec{x}, \Theta),
\end{equation}
where $I$ denotes the photo of the relay wall, depending on the position $\Vec{x}$ of a person and other scene configuration $\Theta$. The scene configuration $\Theta$ mainly includes the appearance of the person, the material of the wall, and the illumination condition. The imaging function $\mathcal{F}$ \textit{compresses} the light field within the hidden region and casts it onto the relay wall.
According to \cref{eq:img_map}, a change in the scene (\eg, a person's position) will cause a change of the \textit{shadow} on the relay wall correspondingly. Mathematically, this change can be formulated as the partial derivative of \cref{eq:img_map}:
\begin{equation} \label{eq:diff_I}
    \frac{\partial I}{\partial \Vec{x}} = \frac{\partial \mathcal{F}(\Vec{x}, \Theta)}{\partial \Vec{x}}.
\end{equation}

Through the guidance of \cref{eq:img_map} and \cref{eq:diff_I}, it is possible to track a person out of LOS by scrutinizing the faint shadow changes. Given a series of discrete observations over time $\{I_0, ..., I_t, ...\}$, \ie, raw frames of a video, NLOS tracking aims to find an inverse imaging function, $\mathcal{F}^{-1}$, which reconstructs the causes $\{x_0, ..., x_t, ...\}$, \ie, the trajectory of the moving objects in real time.

\begin{figure}[t]
  \centering
  \includegraphics[width=\linewidth]{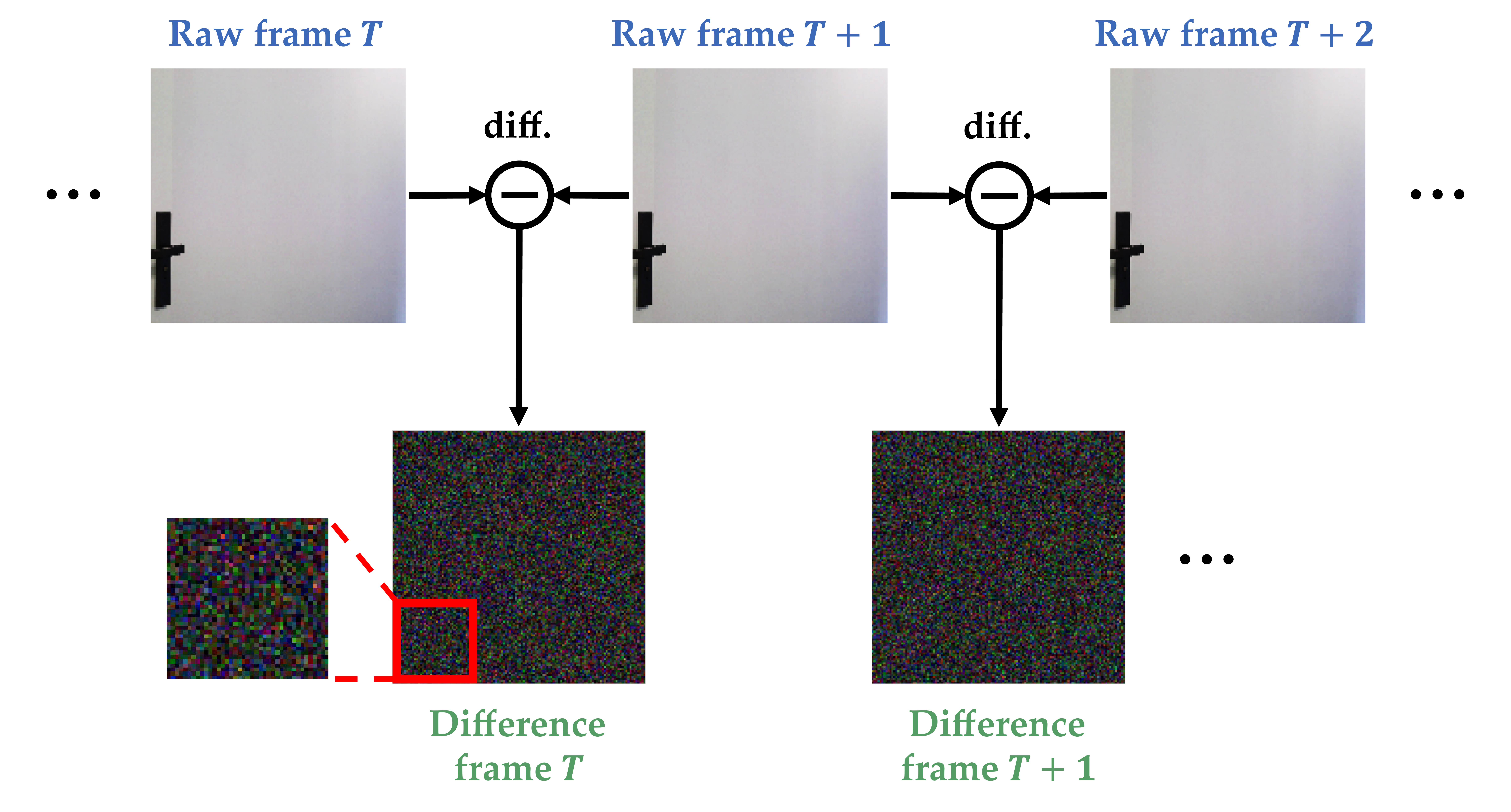}
  \caption{\textbf{Visualization of raw frames and difference frames.} The ``diff." is the abbreviation for ``difference", which means subtracting the previous frame from the current frame. We visualize difference frames after taking absolute value and normalizing to $[0, 1]$ for higher contrast.}
  \label{fig:raw_diff}
\end{figure}

\begin{figure*}[t]
  \centering
  \includegraphics[width=\linewidth]{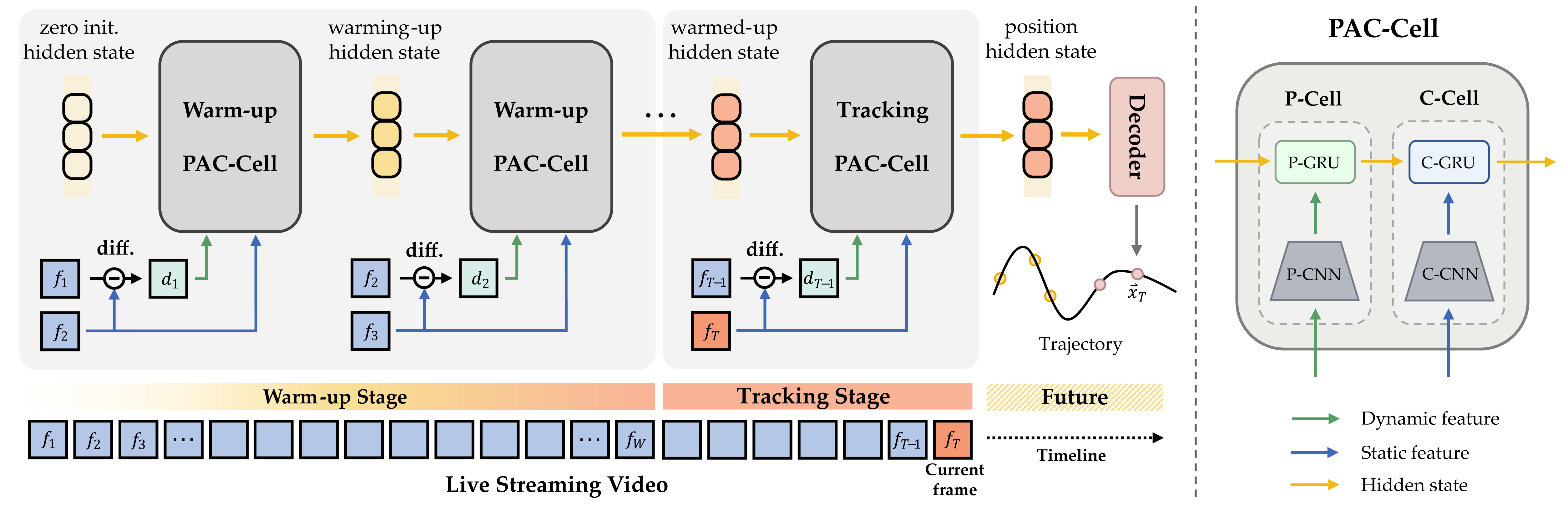}

  \caption{\textbf{Visualization of tracking pipeline with PAC-Net.}
  There are two stages in the whole pipeline, \textit{Warm-up Stage} and \textit{Tracking Stage}. Specifically, given a live streaming video, the Warm-up Stage leverages the first $W$-th frames to ``warm up" the hidden state $\mathbf{h}$ using the Warm-up PAC-Cell. Then the Tracking PAC-Cell infers the trajectory in the Tracking Stage. Both two cells take in dynamic feature (dark green arrow) and static feature (dark blue arrow) and updates the hidden state (light yellow arrow) alternately. Note that PAC-Net can process every incoming frame in an online manner. (Best viewed in color.)
  }
  \label{fig:pipeline}
\end{figure*}

\subsection{Difference Frames} \label{sec:diff_frame}

Most previous works neglect the motion information in tracking tasks, while it can play an important role in guiding the tracking process.
To extract motion information explicitly, we can limit our vision to the vicinity of a single moment $t$. Along with \cref{eq:diff_I}, we can derive the relation between the relay wall's variation $\Delta I_t$ with the person's movement $\Delta \Vec{x_t}$ in the form of finite difference:
\looseness=-1
\begin{equation} \label{eq:delta_I}
\begin{aligned}
\left.\frac{\Delta I}{\Delta \vec{x}} \right|_t &\approx \left. \frac{\partial \mathcal{F}(\vec{x}, \Theta)}{\partial \vec{x}} \right|_t \\
\Longrightarrow I_{t+1} - I_t = \Delta I_t &\approx \left. \frac{\partial \mathcal{F}(\vec{x}, \Theta)}{\partial \vec{x}} \right|_{\vec{x} = \vec{x}_t}  \Delta \vec{x_t} \\
& = \mathcal{G}(\vec{x_t}, \Delta \vec{x_t}, \Theta),
\end{aligned}
\end{equation}
where $\Delta I_t$ is the \textit{difference frame}, which is obtained by subtracting the previous frame from the current frame in the video (\cref{fig:raw_diff}), and $\mathcal{G}$ denotes the imaging function of difference frame. As shown in \cref{eq:delta_I}, the person's movement $\Delta \Vec{x_t}$ directly influences the difference frame $\Delta I_t$. Given a combination of position and motion $\left( \vec{x}, \Delta \vec{x} \right)$, $\mathcal{G}$ maps the tuple to a corresponding difference frame. Therefore, through excavating difference frames, we can further leverage dynamic motion information beyond static positions, which significantly benefits the NLOS tracking task. See \cref{sec:p-net} for more details.

Furthermore, the temporally local nature of difference frames enables them to provide ``clean" motion information. In comparison, although background frame estimation and subtraction~\cite{bouman2017turning, PrafullSharma2021WhatYC, he2022non} can increase SNR, this practice inevitably introduces the information of other time to every single frame, thus making static information ``dirty".

\section{Tracking Method}

When a person is walking in the hidden room, there is a live streaming video coming from the camera shooting at the relay wall. This raw frame stream $\{I_t\}$ incorporates a series of discrete static position information. 
We can readily separate the difference frame stream $\{\Delta I_t\}$ from the raw frame stream, which contains the dynamic motion information instead.
To exploit both motion and position information conveyed by two streams, we propose a concise dual architecture, PAC-Net (\textbf{P}ropagation \textbf{A}nd \textbf{C}alibration \textbf{Net}work). Instead of using two-branch architectures \cite{simonyan2014two} to process two streams separately, PAC-Net integrates the motion continuity prior to its workflow with a specially designed alternating recurrent architecture.

\subsection{PAC-Net} \label{sec:pac-net}

The architecture of PAC-Net and the corresponding real-time tracking pipeline are illustrated in \cref{fig:pipeline}. The design notion of PAC-Net is to process difference and raw stream in an alternate manner, namely \textit{propagate} and \textit{calibrate}. Based on this, we design a dual architecture using the main component, \textit{PAC-Cell}. Within a PAC-Cell, there are two symmetric cells, Propagation-Cell and Calibration-Cell \footnote{Hereafter referred to as P-Cell and C-Cell.}.
Since difference frames convey significant dynamic motion
information as shown in \cref{eq:delta_I}, P-Cell first \textit{propagates} the hidden state between two observations with a difference frame $\Delta I_{T-1}$. Then the newly incoming raw frame $I_T$ introduces absolute position messages, allowing C-Cell to \textit{calibrates} the hidden state to a more accurate one. The whole procedure is as below:
\begin{equation} \label{eq:prop_calib}
\begin{aligned}
&\mathrm{Propagate:}~&\Tilde{\mathbf{h}}_T &= \text{P-Cell} \left( \mathbf{h}_{T-1}, \textcolor{myGreen!90!black}{\Delta I_{T-1}} \right), \\
&\mathrm{Calibrate:}~&\mathbf{h}_T &= \text{C-Cell} \left( \Tilde{\mathbf{h}}_T, \textcolor{myBlue!90!black}{I_T} \right).
\end{aligned}
\end{equation}

In this paper, we use GRU cell \cite{KyunghyunCho2014GRU} as a recurrent cell and ResNet-18 \cite{he2016deep} as a feature extractor, which allow real-time inference with low computational cost. The decoder in \cref{fig:pipeline} is a two-layer Multilayer Perceptron (MLP). In fact, PAC-Net is a framework-level architecture for reconstruction tasks with temporally dense observations. The aforementioned components could be selected accordingly in other tasks.

\begin{figure*}[t]
    \centering
    \includegraphics[width=\linewidth]{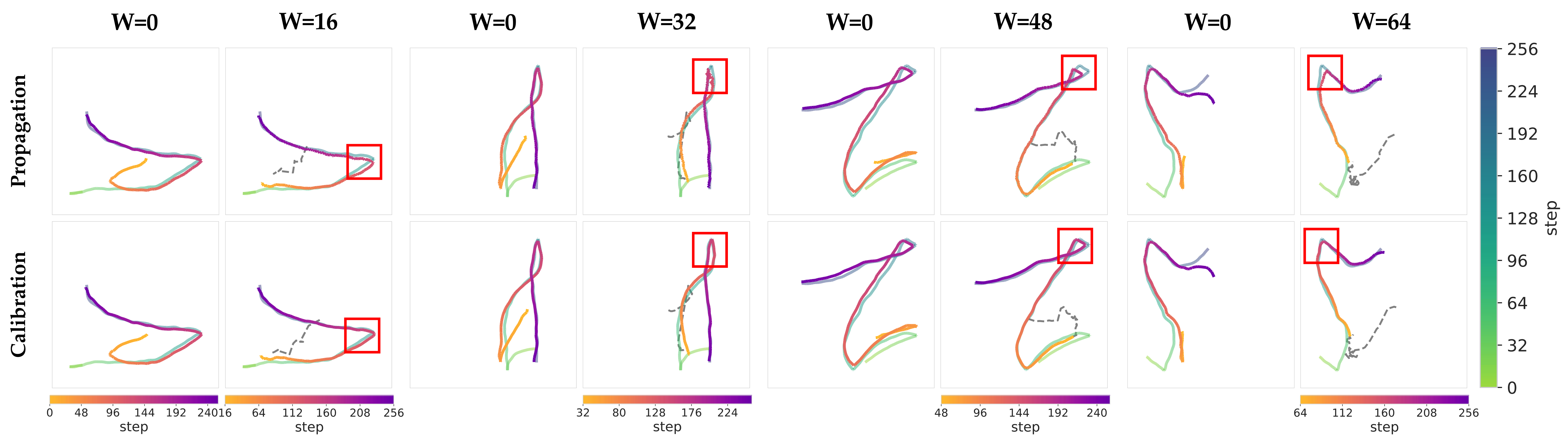}
    \caption{\textbf{Visualization of tracking results with PAC-Net on synthetic data.} Two rows indicate trajectories decoded from P-Cell and C-Cell respectively, called P-trajectory and C-trajectory. Each column indicates different warm-up steps $W$. 
    Grey dashed lines represent the course of the warm-up. 
    Two color gradients for mapping step numbers are applied to ground truths (green) and tracking results respectively. 
    For neat visualization, trajectories are normalized with the room size so that each sub-figure is a square.
    All columns titled $W=0$ share the same color bar as the first column.
    Red squares denote where trajectories are refined by C-Cell. (Best viewed in color.)}
    \label{fig:pc_compare_render}
\end{figure*}

At each time step $T$, the current position $\Vec{x}_T$ can be decoded from the hidden state $\mathbf{h_T}$. So on and so forth, the trajectory of a moving subject can be tracked in real time. Some tracking results are demonstrated in \cref{fig:pc_compare_render} and \cref{fig:model_compare_real}. Note red squares in \cref{fig:pc_compare_render} highlight the difference between trajectories after propagation and calibration. 

Note that the alternating recurrent workflow plays an indispensable role to allow the hidden state $\mathbf{h}$ to propagate stably between observations. Formally, the process of a recurrent neural network (RNN) updating the hidden state could be reformulated as the discretized first-order method for integrating ordinary differential equations (ODEs)~\cite{chen2018neural, de2019gru}:
\begin{equation} \label{eq:rnn_ode}
\begin{aligned}
\frac{d \mathbf{h}(t)}{d t}=f(\mathbf{h}(t), t, \theta) \Longrightarrow  \mathbf{h}_{t} &=\mathbf{h}_{t-1}+f\left(\mathbf{h}_{t-1}, \theta_{t-1}\right) \\
\Longrightarrow \mathbf{h}_{t} &= \Tilde{f} \left( \mathbf{h}_{t-1}, x_{t}\right),
\end{aligned}
\end{equation}
where $x_t$ is the newly incoming observation and $\theta$ represents the parameters. It is natural to consider taking advantage of this intrinsic nature of RNNs to perform tracking. However, our experiments expose the defect of the first-order method (See \cref{sec:results} for more details) that the tracking results either maintain poor continuity or diverge over time. In contrast, the alternating workflow allows PAC-Net to integrate dynamic and static information. This behavior is similar to predictor-corrector methods for solving ODEs, which can ensure numerical convergence via alternately predicting and correcting. Such methods can be formulated as follows:
\looseness=-1
\begin{equation}
\begin{aligned}
    \Tilde{\mathbf{h}}_{t,0} & = \mathbf{h}_{t-1} + f \left(\mathbf{h}_{t-1}, \theta_{t-1}\right), \\
    \Tilde{\mathbf{h}}_{t,n} & = \mathbf{h}_{t-1} + \Tilde{f} \left(\mathbf{h}_{t-1}, \Tilde{\mathbf{h}}_{t,n-1},\theta_{t-1}\right),
\end{aligned}
\end{equation}
where $n=1,2,...,N$ and $N$ denotes the total number of corrections. After predicting with motion information of difference frames, PAC-Net performs a correction with static information of raw frames. This way, PAC-Net can perform tracking coherently and accurately without divergence.

The tracking procedure allows end-to-end training of the model via minimizing the loss with respect to ground truth trajectory. The loss function is composed of position error $loss_x$ and velocity(displacement) error $loss_v$ with an adjustment parameter $\alpha_v$, controlling the weight of $loss_v$. Both of them follow a Mean-Square Error (MSE) fashion:
\looseness=-1
\begin{equation}
\begin{aligned}
    Loss &= loss_x + \alpha_v \cdot loss_v \\
    &= MSE(\{\Tilde{\Vec{x}}_t\}, \{\Vec{x}_t\}) + \alpha_v \cdot MSE(\{\Delta \Tilde{\Vec{x}}_t\}, \{\Delta \Vec{x}_t\}),
\end{aligned}
\end{equation}
where $\alpha_v$ is set to 500 in all experiments to align the orders of magnitude of $loss_x$ and $loss_v$. The velocity loss ensures the model learns a reasonable representation from difference frames and accelerates the convergence of trajectory continuity. Please refer to the supplementary material for more details about model training.

\subsection{Warm-up} \label{sec:warmup}

Since we have no knowledge about the hidden scene before the video stream comes in, we apply a zero-initialization to the hidden state $\mathbf{h}$ in the GRU cell. Although the principle is different, we observe a similar phenomenon as described in keyhole imaging \cite{metzler2020keyhole} -- The tracking trajectories deviate from the ground truth during early steps and gradually converge over time (visualized in columns titled $W=0$ in \cref{fig:pc_compare_render}). This is reasonable because a ``well-defined" hidden state $\mathbf{h}$ doesn't come from nowhere. The convergence over time may indicate gradually gaining knowledge of the unknown room since there are various room sizes and wall materials in our dataset. 

A natural idea is that we could disentangle a few early steps as \textit{Warm-up Stage} from the original tracking procedure. Therefore, we construct two independent PAC-Cells in PAC-Net. The first one is called \textit{Warm-up PAC-Cell}, which is responsible for ``pulling" the hidden state $\mathbf{h}$ from zero initialization to a reasonable distribution, literally, ``warming up" the model. This way, another PAC-Cell, \textit{Tracking PAC-Cell}, could concentrate on accurately tracking by encoding each subsequent frame into a more accurate embedding. In \cref{fig:pc_compare_render} we visualize the course of warm-up with grey dashed lines. The Warm-up Stage could be regarded as an indirect way to find an appropriate initial hidden state $\mathbf{h}$ for the Tracking Stage.
Note if there is a Warm-up Stage, \ie, $W>0$, we only use the inferred trajectory in Tracking Stage to supervise the model training. We compare different steps to perform warm-up and report the results in \cref{sec:results}.

\section{Experiments} \label{sec:experiment}

In this section, we first introduce our proposed dataset, NLOS-Track, then some quality metrics for evaluating the tracking results, and finally verify the effectiveness of our proposed method with experimental studies.

\subsection{NLOS-Track Dataset}

Compared to existing datasets for passive NLOS imaging (\cref{tab:datasets}), NLOS-Track is focused on fitting realistic dynamic scenes. Therefore, rather than including tons of static photographs or using unrealistic objects as tracking targets (\eg, humanoid dolls or cardboard mannequins), NLOS-Track contains more than one thousand clips of videos that shoot at a blank relay wall while a real person is walking around the hidden room. We also render rich realistic scenes that are diverse in room size, walking character, wall material, and lighting condition.

\subsubsection{Real-shot data collection}

\begin{figure}[t]
  \centering
  \begin{subfigure}{\linewidth}
    \includegraphics[width=0.48\linewidth]{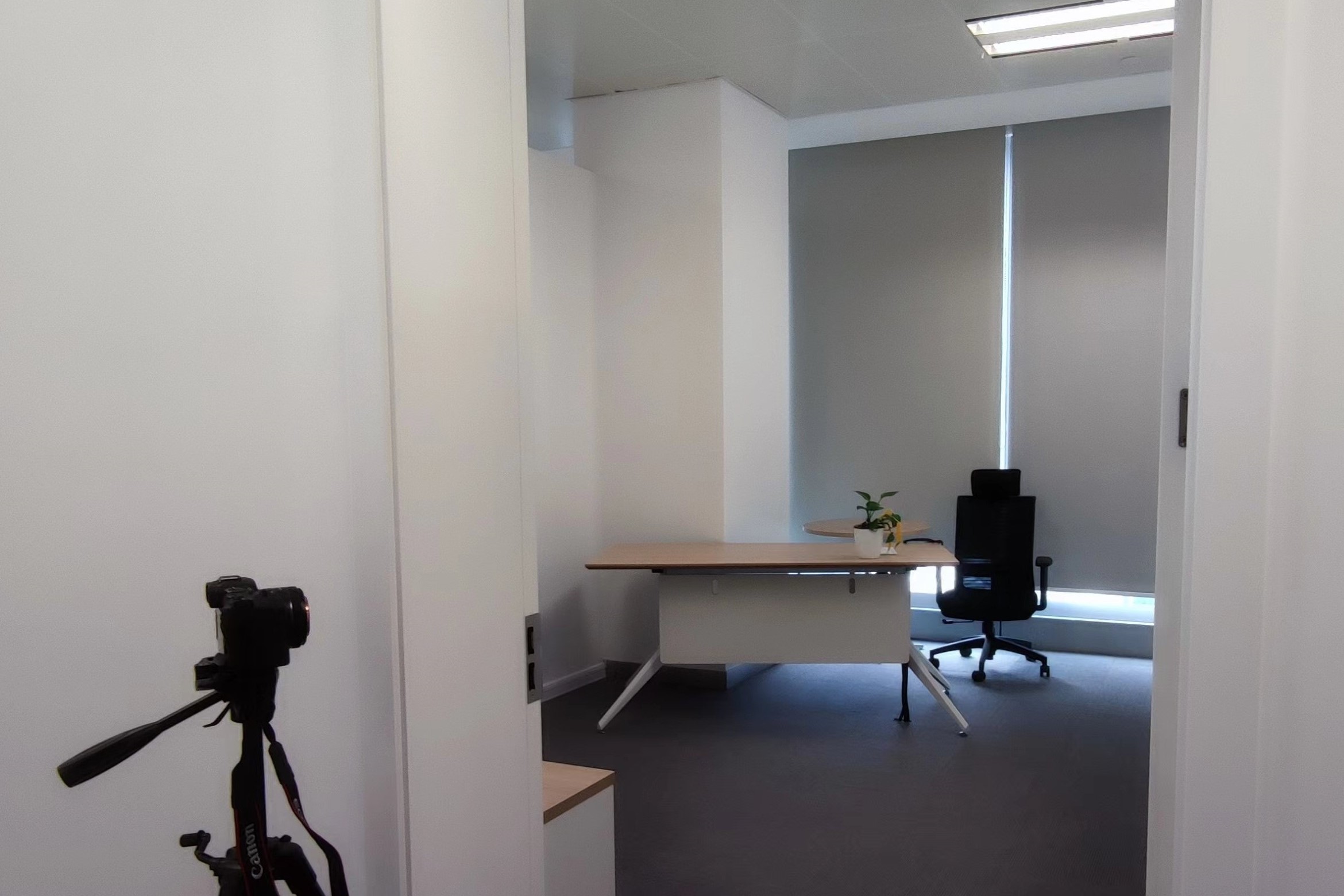}
    \includegraphics[width=0.48\linewidth]{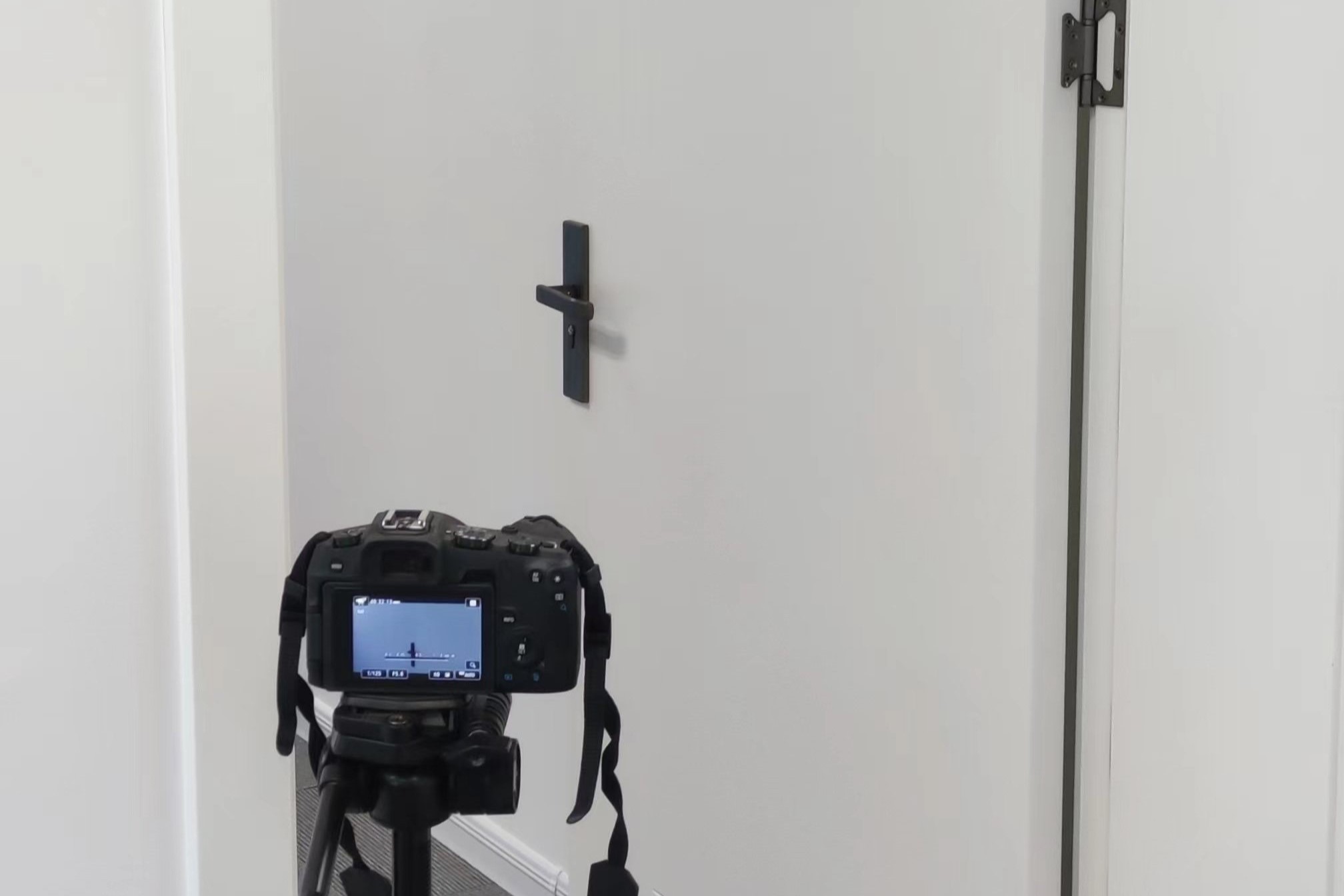}
    \caption{~}
    \label{fig:dataset-real}
  \end{subfigure} \\
  \begin{subfigure}[t]{\linewidth}
    \adjustbox{valign=t}{\includegraphics[width=0.48\linewidth]{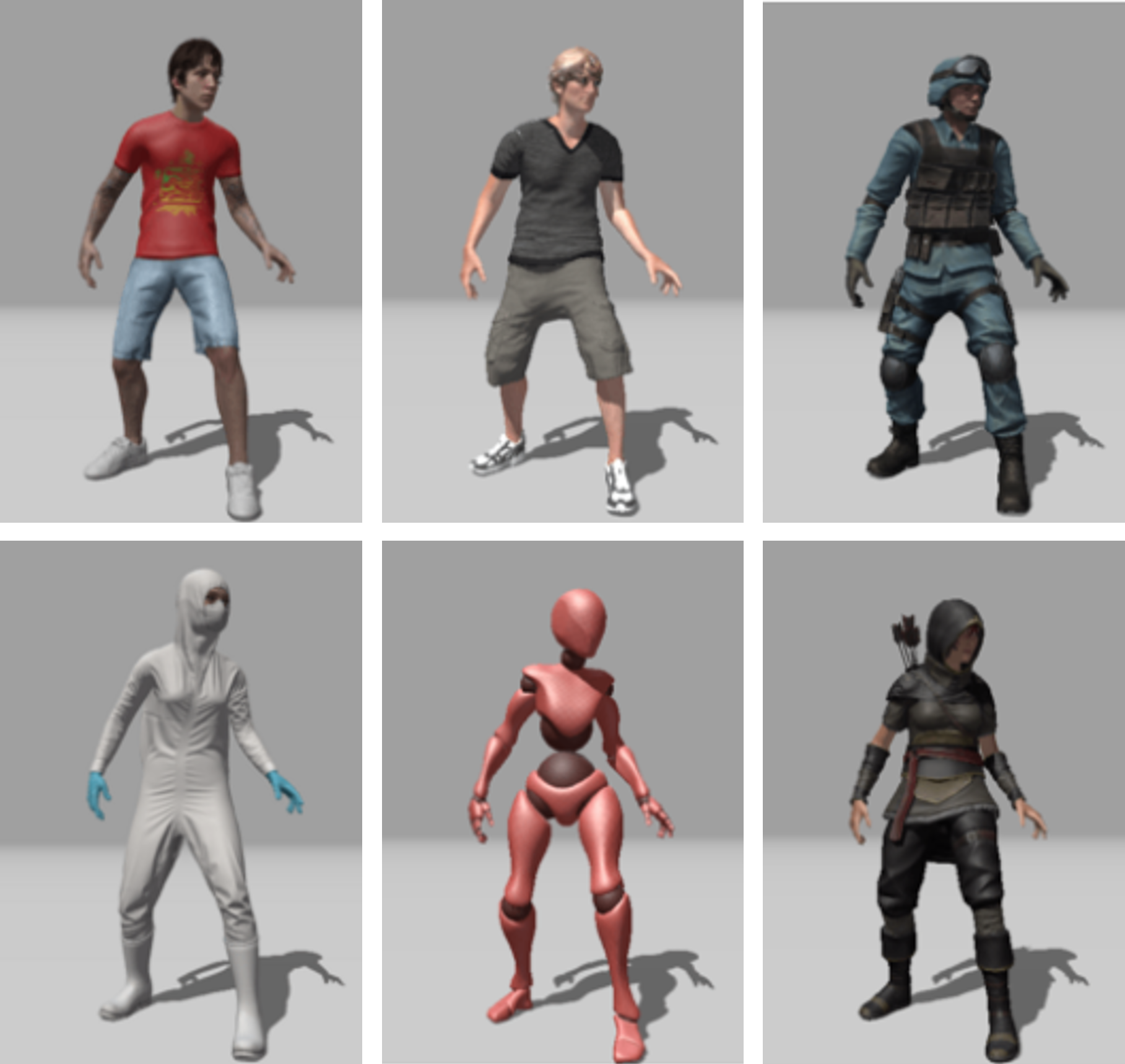}}
    \adjustbox{valign=t}{\includegraphics[width=0.48\linewidth]{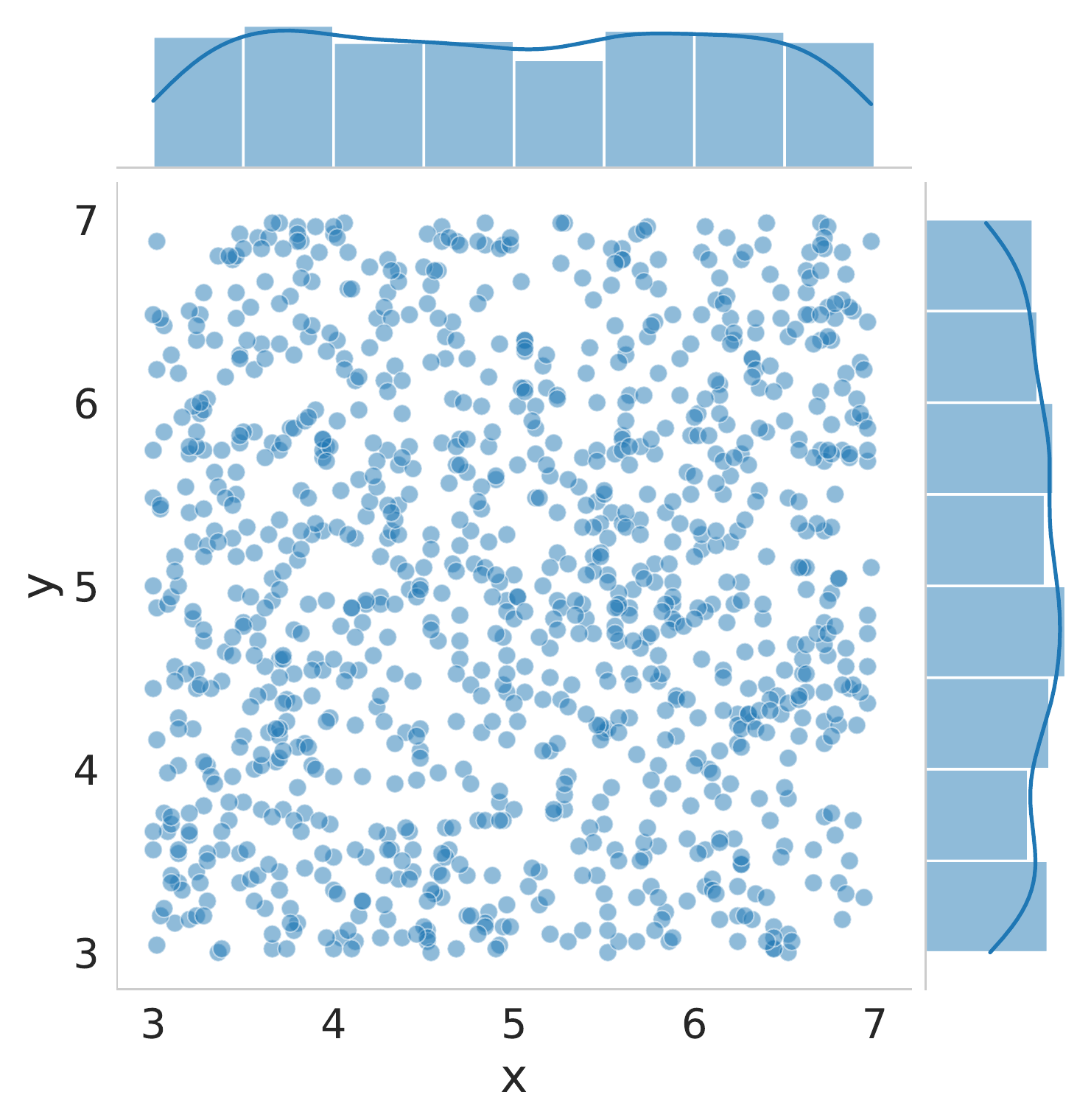}}
    \caption{~}
    \label{fig:dataset-synthetic}
  \end{subfigure}
   \caption{(a) A setup scenario with a camera observing the wall from outside the room. Left: A panorama. Right: View from the camera side. (b) Left: Part of the characters we use in synthetic data. Right: Room size (in meters) distribution and each dot represents a room. }
   \label{fig:dataset}
  \vspace{-10pt}
\end{figure}

We use a consumer-grade micro SLR camera (Canon EOS RP) to capture videos of the relay wall at 25 FPS. To obtain the ground truth of the walking trajectory, we stick a USB camera (HIKVISION E14a) to the ceiling, which records the whole process of people walking from a top view at 25 FPS as well. From the top viewed videos, we are allowed to use Aruco codes to locate the walking person's coordinate frame by frame at a sub-centimeter precision. 
Post-processing of recorded videos includes manually aligning the wall-shooting video stream and coordinate stream and cropping them into video clips of 250 frames.
We ask 3 subjects (each changes 3 different clothes) to walk at various speeds to generalize the dataset.
Please consult supplementary materials for details about the real-shot dataset.
\looseness=-1

\subsubsection{Rendering setup}

In order to reduce the gap between the synthetic data and the photo-realistic data, we use the Cycles render engine in Blender \cite{blender}, which is a physically-based path tracer and provides excellent performance in rendering realistic images. All 3D human-like characters and skeleton models for walking animation are acquired from \href{https://www.mixamo.com/}{Mixamo}, a free animation platform of Adobe. After the character is imported and the walking animation is bound to the character\footnote{Refer to supplementary materials for more details about our random trajectory generation strategy.}, we use the Cycles render engine to conduct steady-state rendering of the relay wall frame by frame. All video clips have 320 frames each, at 30 FPS, and a resolution of $256 \times 256$ pixels. The videos are firstly rendered into \verb|.png| sequences with 8-bit RGB color depth and then pre-processed into \verb|.npy| files for IO efficiency.

On an NVIDIA A100 graphics card, the rendering speed is about 0.8 seconds per frame at a resolution of $256 \times 256$ pixels. In total, we spend about 70 hours rendering the full synthetic dataset of 1000 video clips.

\begin{table}
  \centering
  \scalebox{0.7}{
  \begin{tabular}{@{}lcccc@{}}
    \toprule
    Dataset & Modal & Data Source  & Setup & Size \\
    \midrule
    \multirow{2}*{Platform\cite{klein2018quantitative}} & \multirow{2}*{Transient}  & \multirow{2}*{Synthetic}  & Static \&     & \multirow{2}*{/} \\
    ~                                                   & ~                         & ~                         & Dynamic \\
    Z-NLOS\cite{galindo2019dataset}                     & Transient                 & Synthetic                 & Static        & 300 measurements  \\
    NLOS-Passive\cite{geng2022passive}                  & Steady                    & Real-shot                 & Static        & 3,200,000 images\\
    NLOS-ES\cite{Wang2022event}                         & Event                     & Real-shot                 & Dynamic       & 4,180 images\\
    \midrule
    \specialrule{0em}{1pt}{1pt}
    \midrule
    NLOS-Track                                          & \multirow{2}*{Steady}     & Real-shot                 & \multirow{2}*{Dynamic} & $\sim$1,500 videos\\
    (Ours)                                              & ~                         & \& Synthetic              & ~             & (445,000 frames)\\
    \bottomrule
  \end{tabular}
  }
  \caption{\textbf{A brief summary of existing NLOS datasets.} Transient and Steady under the Modal column denote time-resolved transient-state and conventional RGB steady-state respectively. ``/" denotes difficulty to report the data size due to mixed data types.}
  \label{tab:datasets}
  \vspace{-10pt}
\end{table}

\vspace{-5pt}
\subsubsection{Other considerations on generalization} \label{sec:dataset_generalization}

We randomize several settings before rendering each clip of the video for generalization purposes. The room sizes (in meters) are sampled from a uniform distribution $U(3, 7)$, as shown in \cref{fig:dataset-synthetic}. Besides, the character is randomly selected among more than 20 characters with a variety of outfits. We also change the position and luminosity of light sources and the camera position. Four different floor styles are applied randomly. The texture and roughness of the photographed wall are randomized as well. All these settings are recorded in a \verb|.yaml| file to guarantee reproducibility.

Besides, we simulate the real-world noise in synthetic data to make them more realistic. We first compute the mean and standard deviation of real data and synthetic data at the pixel level and then aligned their statistics by adding noise to the rendering results.

\subsection{Metrics}
As suggested by Klein \etal \cite{klein2018quantitative}, the root-mean-square (RMS) error could be used to evaluate the position inference and tracking quality by directly computing the Euclidean distance between tracking and GT trajectory. Since RMS error is positively correlated with the loss function (MSE) we use, we refer to other three metrics for a comprehensive evaluation -- area between two curves (Area) \cite{jekel2019similarity}, dynamic time warping (DTW) \cite{senin2008dynamic}, and partial curve mapping (PCM) \cite{witowski2012parameter}. All three metrics are based on similarity measures between two curves. To specify, metrics reported in \cref{tab:ablation} are only computed in the Tracking Stage and are normalized by trajectory length for fairness.

\subsection{Results}\label{sec:results}

\begin{figure*}[t]
    \centering
    \includegraphics[width=\linewidth]{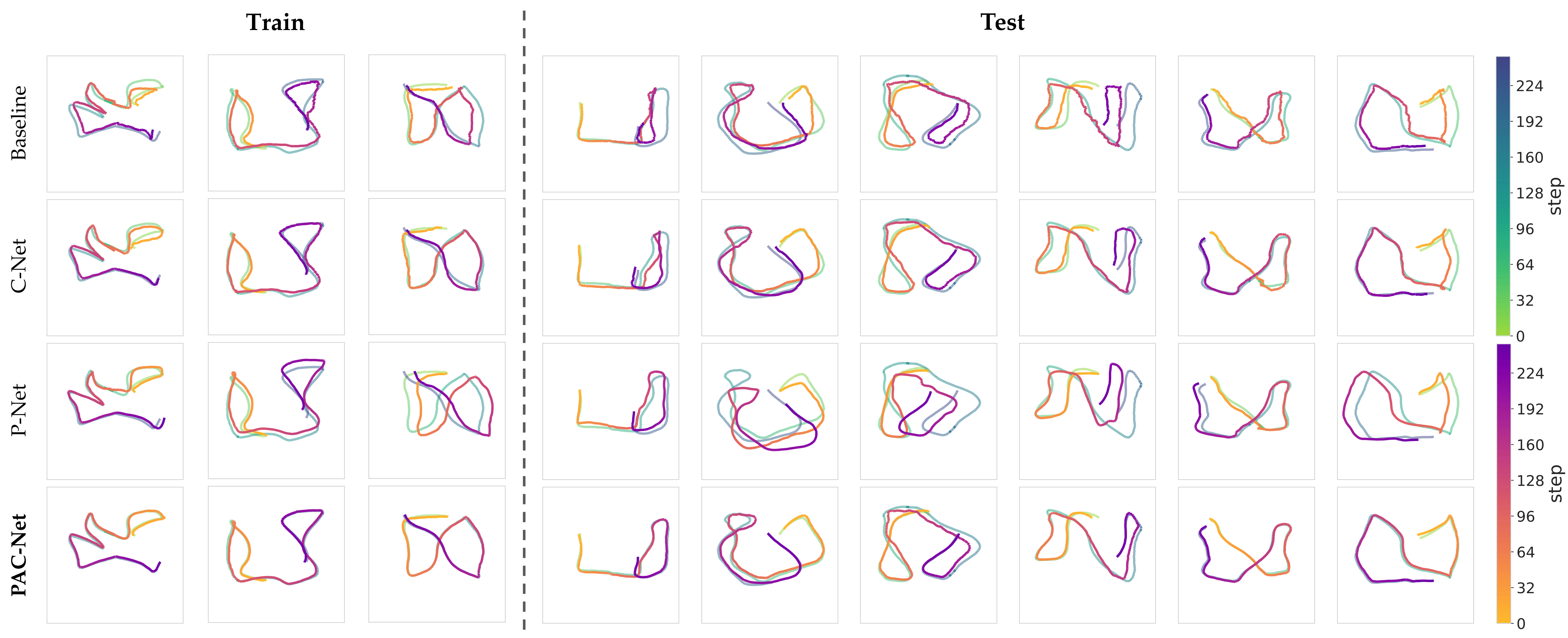}
    \caption{\textbf{Visualization of tracking results with different models on the real-shot dataset.} Rows indicate different methods and columns indicate different trajectories sampled from the Train and Test set. Color gradients share the same meaning as in \cref{fig:pc_compare_render}.
     (Best viewed in color.)\looseness=-1
    }
    \label{fig:model_compare_real}
\end{figure*}

In this section, we describe the experimental results of our proposed method, along with the baseline method and ablation study. It is worth mentioning that we don't compare different choices of components (\eg, various feature extractors, different RNN cells, dimension of decoder) in experiments because our goal is to demonstrate the effectiveness of PAC-Net's architecture, not to improve the performance.
\looseness=-1

\noindent \textbf{PAC-Net.}~
To verify our design notion of PAC-Net, we illustrate trajectory output by P-Cell and C-Cell separately in two rows of \cref{fig:pc_compare_render}, which are called P-trajectory and C-trajectory. 
P-Cell first uses motion information conveyed by difference frames to step forward, which forms P-trajectory. Then under collaboration with C-Cell, P-trajectory is refined into the more accurate C-trajectory with position information introduced by raw frames. In \cref{fig:pc_compare_render} we use red squares to denote sudden changes of velocity, where C-Cell plays a clear calibrating role. Numerical metrics reported in \cref{tab:ablation} show that PAC-Net not only achieves a centimeter-level precision in Tracking Stage but also demonstrates robustness on both real-shot data and synthetic data, thus outperforming other methods. 

To validate the effectiveness of our method, we construct two degenerated models by removing P-Cell or C-Cell from PAC-Cell, called C-Net and P-Net respectively. Each of the two degenerated models only takes in raw frames or difference frames. We also evaluate a CNN-based baseline model, fed with raw frames. Note we use a ResNet-34 and a two-layer GRU in degenerated models and a ResNet-50 as CNN baseline to align the parameter number.

\begin{figure}[t]
\vspace{-15pt}
    \centering
    \begin{subfigure}{0.32\linewidth}
        \includegraphics[width=\linewidth]{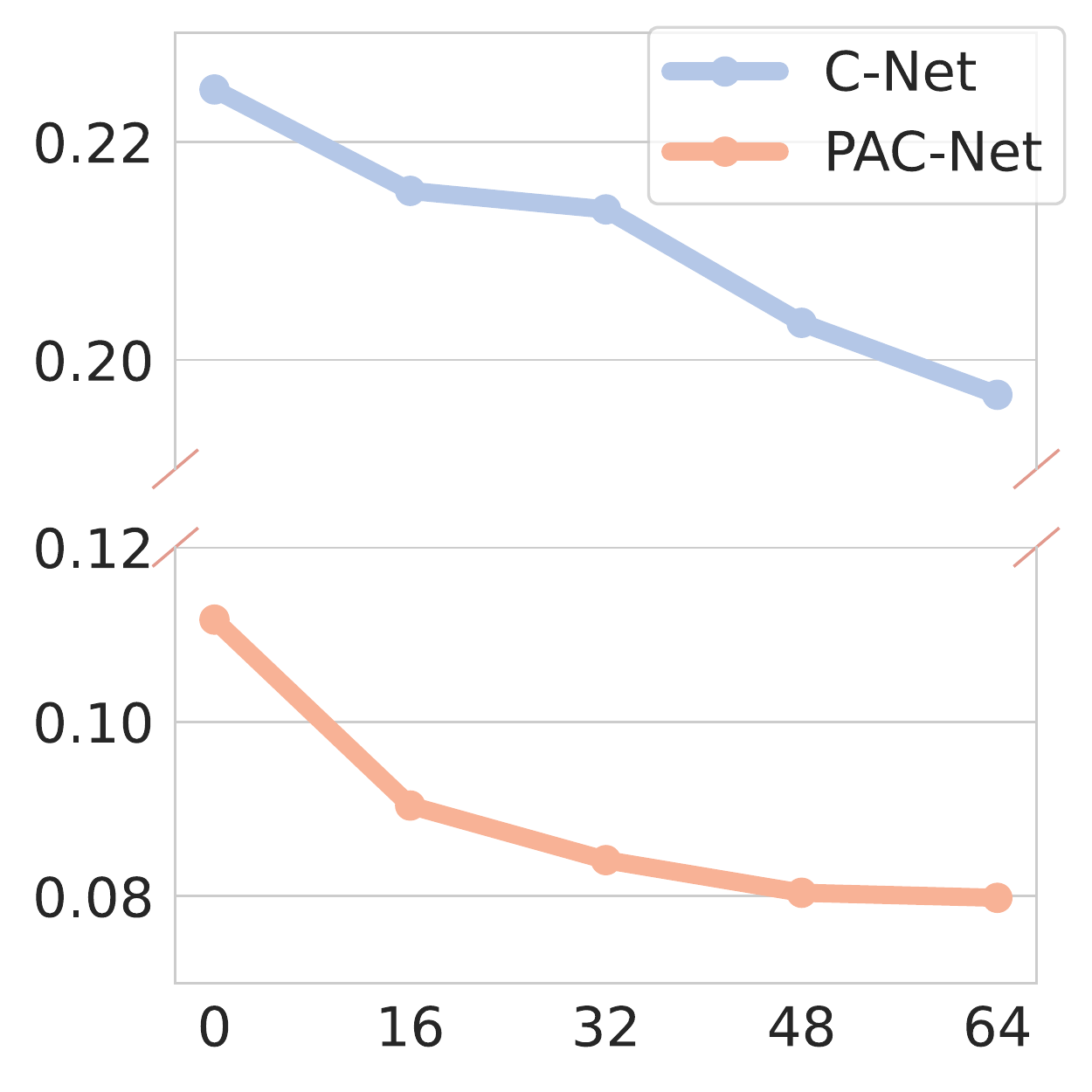}
        \caption{Area ($\downarrow$)}
        \label{fig:diff_warmup-a}
    \end{subfigure}
    \begin{subfigure}{0.32\linewidth}
        \includegraphics[width=\linewidth]{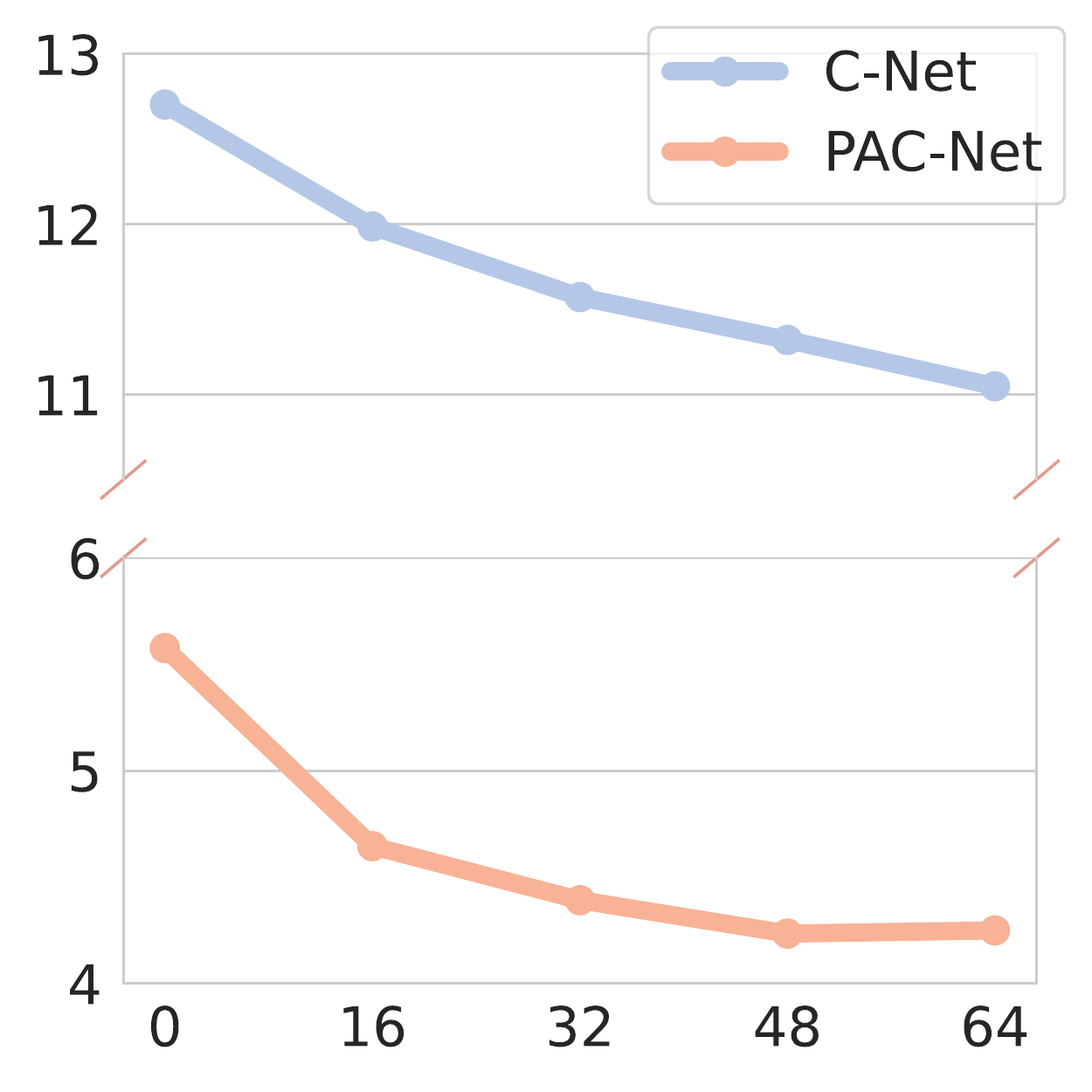}
        \caption{DTW ($\downarrow$)}
        \label{fig:diff_warmup-b}
    \end{subfigure}
    \begin{subfigure}{0.32\linewidth}
        \includegraphics[width=\linewidth]{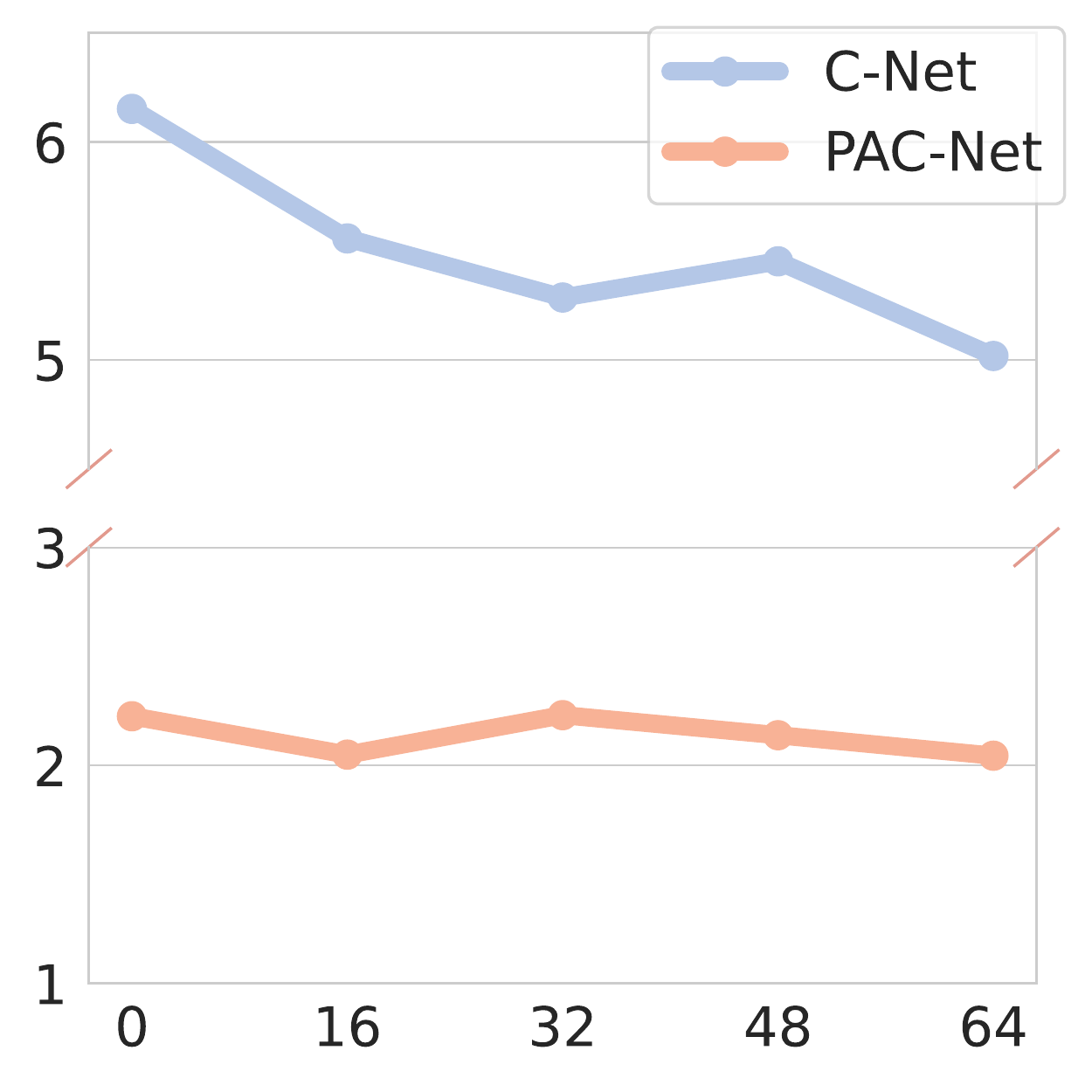}
        \caption{PCM ($\downarrow$)}
        \label{fig:diff_warmup-c}
    \end{subfigure}
    \caption{\textbf{Comparison of different warm-up steps on synthetic data.} Both C-Net and PAC-Net have better tracking performance as warm-up steps increase with a gradual saturation.}
    \label{fig:diff_warmup}
    \vspace{-15pt}
\end{figure}

\noindent \textbf{Baseline.}~ There are obvious jitters in reconstructed trajectories using vanilla CNN, which is shown in the first row of \cref{fig:model_compare_real}. Poor statistic metrics reported in \cref{tab:ablation} prove the necessity of taking advantage of recurrent structure to maintain the trajectory continuity.

\noindent \textbf{C-Net (without dynamic motion information).}~ \label{sec:c-net}
Although C-Net stabilizes the trajectory to some extent via recurrent structure, as shown in the second row of \cref{fig:model_compare_real}, there are still jitters and discontinuities in trajectories. This observation indicates that it is not sufficient to only use position information in passive NLOS tracking.

\noindent \textbf{P-Net (without static position information).}~ \label{sec:p-net}
P-Net has the same structure as C-Net, but takes as input difference frames and reconstructs displacement (velocity) between observations instead. 
To compute metrics using motion information independently, we provide the true initial position so P-Net can accumulate velocities into a complete trajectory. 
As shown in the third row of \cref{fig:model_compare_real}, P-Net is capable of maintaining better smoothness and stability than C-Net, but there is an obvious overall translation with respect to ground truth.

\begin{table*}[t]
\vspace{5pt}
\centering
\begin{tabular}{@{}cccccccc@{}}
    \toprule
    \multirow{2}*{Datasets} & \multicolumn{2}{c}{Methods} & \multicolumn{5}{c}{Metrics} \\
    \cmidrule(lr){2-3} \cmidrule(l){4-8}
    ~ & Type & Model & RMS$_x(\times 10^{-2})$ & RMS$_v(\times 10^{-3})$ & Area~($\downarrow$) & DTW~($\downarrow$) & PCM~($\downarrow$) \\
    \cmidrule{1-8}
    \multirow{6}*{Real-shot}
    ~ & CNN                         & ResNet-50         & 2.60           & 2.94            & 0.01184           & 0.8559            & 0.8171 \\
    \specialrule{0em}{2pt}{1pt}
    ~ & \multirow{3}*{ConvRNN}      & C-Net             & 1.65           & 2.24            & 0.00644           & 0.4617            & 0.4819 \\
    ~ & ~                           & C-Net + Warm-up   & 1.55           & 2.17            & 0.00601           & 0.4371            & 0.4392 \\
    ~ & ~                           & P-Net             & 4.03           & 2.87            & 0.01137           & 0.9727            & 1.211 \\
    \specialrule{0em}{2pt}{1pt}
    ~ & \multirow{2}*{\textbf{Ours}}& PAC-Net           & 1.46           & \textbf{1.17}   & \textbf{0.004347} & 0.3367            & 0.3313 \\
    ~ & ~                           & PAC-Net + Warm-up & \textbf{1.37}  & 1.28            & 0.004388          & \textbf{0.3348}   & \textbf{0.3027} \\
    \midrule
    \specialrule{0em}{1pt}{1pt}
    \midrule
    \multirow{6}*{Synthetic}
    ~ & CNN                         & ResNet-50         & 22.0             & 3.90             & 0.2807            & 21.891            & 316.806 \\
    \specialrule{0em}{2pt}{1pt}
    ~ & \multirow{3}*{ConvRNN}      & C-Net             & 15.5             & 3.31             & 0.2248            & 12.703            & 6.151 \\
    ~ & ~                           & C-Net + Warm-up   & 15.1             & 3.16            & 0.2138            & 11.572            & 5.286 \\
    ~ & ~                           & P-Net             & 11.3             & 2.17            & 0.1251            & 5.774             & 2.826 \\
    \specialrule{0em}{2pt}{1pt}
    ~ & \multirow{2}*{\textbf{Ours}}& PAC-Net           & 9.70            & 2.21            & 0.1117            & 5.576             & \textbf{2.225} \\
    ~ & ~                           & PAC-Net + Warm-up & \textbf{8.78}   & \textbf{1.79}   & \textbf{0.08413}  & \textbf{4.391}    & 2.268 \\
    \bottomrule
\end{tabular}
\caption{\textbf{Evaluation metrics of different models on different datasets.} In metric columns with a down arrow $\downarrow$, the lower metric denotes the better performance of the corresponding model. 
Note that P-Net doesn't have a warm-up stage because the initial position is given. 
Models with ``+ Warm-up" use warm-up steps $W=32$. 
\textbf{Bold} denotes the best-performing models on each metric.}
\label{tab:ablation}
\vspace{-5pt}
\end{table*}

\noindent \textbf{Different Warm-up Steps.}~ 
To evaluate the contribution of the Warm-up Stage, we evaluate PAC-Net and C-Net with different warm-up steps. Note that a longer Warm-up Stage doesn't introduce extra computational cost during inference, but only delays the beginning of the Tracking Stage.
In \cref{fig:pc_compare_render} we demonstrate some tracking results with different warm-up steps. With more steps to warm up, the trajectory will be more accurate when the tracking stage begins. We also observe a gradual saturation of the model's performance as warm-up steps increase. This trend can be clearly seen with statistic metrics shown in \cref{fig:diff_warmup} and warm-up course (grey dashed lines) demonstrated in \cref{fig:pc_compare_render}. With a warm-up step $W \approx 48$, the Warm-up PAC-Cell has taken in sufficient frames to warm up the hidden state $\mathbf{h}$ thus providing a reasonable initialization for the following Tracking Stage.
\looseness=-1

\noindent \textbf{Real-time Inference.}~
We run our model on a laptop with 8-core AMD Ryzen 7 5800H CPU and an Nvidia GeForce RTX 3060 laptop GPU. We achieve an inference speed of approximately 900 frames per second and therefore adequate for real-time inference. The single-scene inference speed on one card of NVIDIA A100 is about 5000 FPS.
\looseness=-1

\section{Limitations and Future Work}

Now our work is limited to 2D indoor tracking. However, our proposed method is also compatible with 3D tracking in the wild, which remains not validated due to the difficulty of collecting and labeling 3D data. It is prospective to extend our method to 3D scenes, which will meet more diversified real needs in fields like autonomous driving and security.
Apart from that, our pipeline is now applied to single-object tracking tasks. It is natural to consider extending our method to multi-object tracking. The analysis of this problem will be difficult because the complexity of motion information doesn't accumulate linearly with the number of objects.
Besides, developing self-supervised or semi-supervised techniques on NLOS tracking is also a promising perspective, which helps the generalization of models.
\looseness=-1

\section{Conclusion}

In this paper, we formally propose and formulate the task of real-time passive NLOS tracking.
We introduce difference frame which conveys clean motion information and helps to reconstruct a continuous and smooth trajectory. 
In addition, we propose a novel and concise network architecture, PAC-Net, which is capable of maintaining good continuity and stability via processing raw frames and difference frames alternately.
Note that PAC-Net is a framework-level architecture, and thus can be applied to other reconstruction tasks with dense observation in time series.
We mitigate the dilemma of initialization of a recurrent-fashion network to some extent by disentangling an independent Warm-up Stage.
To evaluate our proposed method and facilitate data-driven techniques, we establish the first passive NLOS tracking dataset, NLOS-Track, which contains about 1500 videos of realistic scenes. Both real-shot data and synthetic data are included to generalize the dataset.

\noindent{\textbf{Acknowledgement.}}
This research is supported in part by
Shanghai AI Laboratory, the Natural Science Basic Research Program of Shaanxi under Grant 2021JQ-204 and the National Natural Science Foundation of China under Grant 62106183 and 62106182.

{\small
\bibliographystyle{ieee_fullname}
\bibliography{bib_main}
}

\end{document}


\title{Supplementary Material for\\Propagate And Calibrate: Real-time Passive Non-line-of-sight Tracking}

\author{}

\maketitle

    

\section{NLOS-Track Dataset}

\subsection{Real-shot Data}

\subsubsection{Video collection and position calibration}
To obtain paired wall-shooting videos and ground truth trajectories, we record the relay wall with one camera and stick another camera on the ceiling to have a corresponding top-viewed video at the same time.
Subjects walk around the hidden room with Aruco code on head so that we can track him with the top-viewed video. To improve the accuracy and robustness of tracking, we put four Aruco codes on the four corners of a hard and flat board. 
We use the Aruco API provided in OpenCV~\cite{bradski2000opencv} to obtain the 3D pose of Aruco codes with respect to ceiling camera. 
In each frame of the top-viewed video, only when all four codes are detected, we take the average of four codes' coordinates as the character's position coordinate. Then translation and rotation are applied to transform the character's coordinate from the camera system to the world system.

\subsubsection{Stream alignment and cropping}
We use two pulse signals that only last for a short moment to mark the beginning and the ending time point of a single clip. Two signals are designed to be visible by exposing them to both two cameras.
Since the time intervals of the two signals in the two videos are equal, we are allowed to manually align the two videos at a frame-level precision by aligning the beginning time point and the ending time point in the timeline. 
\looseness=-1

\subsubsection{Data cleaning}
Although we use four codes to improve the tracking robustness, there are still some point-sequences that we fail to track due to jitters and blurs of pictures. 
Assume that the missing points in one sequence are $\{\Vec{p}_i\}$, we use a linear interpolation to complete them:
\begin{equation}
\begin{aligned}
    \Vec{p}_{i} &\approx  \frac{N-i}{N}\Vec{p}_0 + \frac{i}{N}\Vec{p}_N \\
    &= \Vec{p}_0 + \frac{i}{N}\left( \Vec{p}_N - \Vec{p}_0 \right),\quad i=1,2,...,N-1,
\end{aligned}
\end{equation}
where $i$ is the index of the missing point. $\Vec{p}_0$ denotes the recorded point previous to the missing sequence while $\Vec{p}_N$ denotes the subsequent recorded point. To perform this interpolation, we make a reasonable assumption that there are no sharp turns or changes in speed within $N$ frames. In addition, we set a threshold of $N \le 10$ when conducting the interpolation, which reinforces this assumption so that we won't introduce unbearable errors during data cleaning.

After completing missing points and excluding clips that can not be completed, we crop the remaining paired videos and ground truth trajectories into many 250-frame clips. The videos are save as \verb|.npy| files and trajectories are saved as \verb|.mat| files, along with corresponding room size.

\subsection{Random Trajectory Generation}

To simulate the realistic situation of people walking in the room, we use a heuristic algorithm for generating near-real continuous trajectories frame by frame. Given the room size, we first select a random position $\Vec{p}_0=(p^x_0, p^y_0)$\footnote{Here we use $p$ to denote position instead of $x$, which is distinguished from $x$ used to denote direction.} within the room as the initial position and a random vector $\Vec{v}_0=(v^x_0, v^y_0)$ as the initial walking direction. Considering the effect of inertia and momentum in real world, in each subsequent frame we make a slight and random change $\Delta \Vec{v}_{t}$ to the current walking direction as that of the next step:
\begin{equation}
    \Vec{v}_{t+1} = \Vec{v}_{t} + \tau \cdot \Delta \Vec{v}_{t} ~,~ t=0,1,...,
\end{equation}
where $\tau$ is an adjustable parameter called \textit{turning rate}.
By setting $\tau$ properly, one can control the curvature degree when generating trajectories. Specifically, we set the turning rate $\tau$ to 0.15 in our dataset because this value guarantees a reasonable continuity and rotation magnitude of generated trajectories.
In each frame, the character takes a step forward in the new direction $\Vec{v}_{t+1}$ and a continuous trajectory can be generated:
\begin{equation}
    \Vec{p}_{t+1} = \Vec{p}_{t} + \Vec{v}_{t} ~,~ t=0,1,...
\end{equation}
Note that the distance (in meters) of each frame step is not fixed, but is randomly sampled with a uniform distribution $U(0.03, 0.04)$ for the sake of being realistic.

In addition, we ensured that the characters did not cast any direct shadow on the wall that would be visible to the naked eye while generating the trajectory.

\subsection{Data Format in Dataset}

We store all video clips in \verb|.npy| format after cropping them to squares and resizing them to a dimension of $T \times C \times H \times W$, where $C=3$ is the number of channels and $H=W=128$ is the spatial dimension. $T=250$ in real-shot data while $T=320$ in synthetic data.

\section{PAC-Net}

\subsection{Model Details}

We use PyTorch~\cite{Paszke_PyTorch_An_Imperative_2019} to build and train our neural networks. Besides ResNets~\cite{he2016deep}, there are only two other basic components
in our model, which are GRU cells \cite{KyunghyunCho2014GRU} and a self-designed MLP decoder. 
The vector dimension of extracted features and hidden state $\mathbf{h}$ are both 128.
Before loading the pre-trained weights of ResNets, we substitute the last linear layer in ResNets with another one with an output dimension of 128. 
The two-layer MLP decoder takes in the hidden state $\mathbf{h}$ as input. The 64-dimensional intermediate vector is activated by a ReLU, followed by a final linear layer that outputs the 2-dimensional coordinate.

\subsection{Model Training}

During offline training, we use an equivalent implementation of networks to improve the training efficiency. Instead of extracting feature frame-by-frame, we use two ResNets in PAC-Cell to extract features from raw frames and difference frames all at once after loading a video clip. Then static features (from raw frames) and dynamic features (from difference frames) are fed to P-GRU and C-GRU alternatively.
\looseness=-1

For all networks, we train models for 70 epochs using AdamW~\cite{loshchilov2017decoupled} optimizer with a weight decay of 2e-3. The learning rate is set to 3e-4 and follows the cosine annealing schedule~\cite{loshchilov2016sgdr}. The batch size is set to 32.

During training, we load a random $T'$-frame clip from the whole video clip of $T$ frames, where $T'\le T$. This practice further enhances the diversity of datasets.

\section{More Analysis}

\subsection{Light sources}

We set all three light sources on the ceiling (please refer to Fig.1 in the main paper) so the useful information on the wall mainly comes from the diffuse reflection rather than obvious shadows (please refer to Sec.3 in the main paper). Since the tracked person is always moving, the position of lights relative to the person is not fixed.

\begin{figure}[t]
    \vspace{-15pt}
    \centering
    \includegraphics[width=\linewidth]{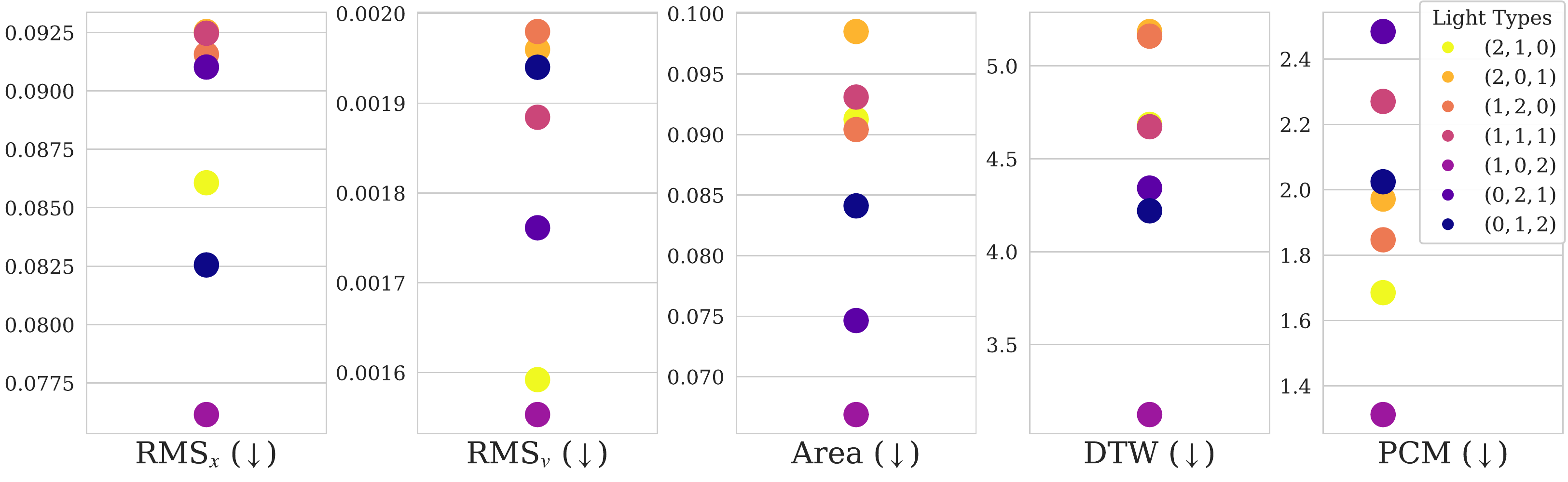}
    \vspace{-18pt}
    \caption{\textbf{How light sources affect metrics.} Light type of $(1,0,2)$ means 1 point light, 0 spot light and 2 area lights are in the room.}
    \label{fig:light_analysis}
    \vspace{-10pt}
\end{figure}

In the synthetic dataset we vary the \textit{type(point, spot, and area), position, rotation}, and \textit{power} of three light sources. With so many factors influencing the light condition, it is a ponderous task to conduct a fair and comprehensive investigation. In \cref{fig:light_analysis}, we preliminarily show how metrics vary with different light type combinations. We can see that with 2 area lights, we have relatively good tracking results on average because more area light makes the room brighter, thus can provide sufficient diffuse signal cast on the relay wall for NLOS tracking. We found there is a coarsely positive correlation between light source area and performance.

\subsection{Warm-up}

When facing a variety of room settings in synthetic data, the Warm-up stage plays an important role in ``adapting" to the current room (please refer to the grey dashed lines in Fig.4 in the main paper). In contrast, due to the relatively limited real-shot scenes, the potential of the Warm-up stage cannot be fully demonstrated.
We conduct an extra comparison test on a small synthetic dataset with only two room sizes. And we observe only minor differences between w/ and w/o Warm-up (\cref{tab:warm-up}). Thus we conclude that Warm-up requires a diverse dataset to work.

\begin{table}[ht]
    \centering
    \vspace{-5pt}
    \scalebox{0.8}{
    \begin{tabular}{cccccc}
        \toprule
        Model & RMS$_x$ & RMS$_v$\small{$(\times 10^{-3})$} & Area & DTW & PCM  \\
        \hline
        w/o Warm-up & \textbf{0.1103} & 2.50 & 0.0776 & 1.698 & \textbf{2.613} \\
        w/ Warm-up & 0.1105 & \textbf{2.03} & \textbf{0.0715} & \textbf{1.589} & 3.429 \\
        \bottomrule
    \end{tabular}
    }
    \vspace{-5pt}
    \caption{\textbf{Comparison test on warm-up.}}
    \label{tab:warm-up}
    \vspace{-10pt}
\end{table}

\section{More Visualisations}

To demonstrate the test results in real scenes in an intuitive way, we attach two demo videos in supplementary materials\footnote{Demos can also be found on the \href{https://againstentropy.github.io/NLOS-Track/}{project website}.}. In each video, we present the equivalent real-time reconstructed trajectory, along with the corresponding raw stream and difference stream. We use red squares to highlight when and where the faint variation in the relay wall can be detected by naked eyes with difference frames.

\newpage
{\small
\bibliographystyle{ieee_fullname}
\bibliography{bib_supp}
}